%% file: main.tex
\documentclass[11pt, a4paper, logo, onecolumn, nonumbering]{googledeepmind}

\pdftrailerid{redacted}
\usepackage{xspace}
\input{preamble.tex}
\title{Reinforced Attention Learning} 
\usepackage[numbers, sort&compress]{natbib}
\correspondingauthor{bzhli@ucdavis.edu}
\def\modelname{\texttt{\textbf{RAL}}\xspace}
\def\modelfullname{\texttt{\textbf{Reinforced Attention Learning}}\xspace}
\definecolor{googleblue}{HTML}{4285F4}
\definecolor{googlered}{HTML}{EA4335}
\definecolor{googlegreen}{HTML}{34A853}
\definecolor{googleyellow}{HTML}{FBBC04}
\reportnumber{} %
\usepackage{amsmath}
\usepackage{amssymb}
\usepackage{mathtools}
\usepackage{amsthm}
\usepackage{booktabs}
\usepackage{multirow}
\usepackage{tcolorbox} 
\tcbuselibrary{most}   
\usepackage{hyperref}
\usepackage{cleveref}  
\usepackage{float} 
\author[1]{Bangzheng~Li\textsuperscript{*}}
\author[3]{Chen~Qu}
\author[4]{Jianmo~Ni}
\author[3]{Ian~Miao}
\author[3]{Liu~Yang}
\author[2]{Xingyu~Fu}
\author[1]{Muhao~Chen}
\author[4]{Derek~Zhiyuan~Cheng}
\author[4]{Zhong~Cheng}

\affil[1]{UC Davis}
\affil[2]{Princeton University}
\affil[3]{Google}
\affil[4]{Google DeepMind}

\input{sections/00_abstract}

\begin{document}

\newpage
\maketitle

\input{sections/01_intro}

\input{sections/02_related_works}
\input{sections/03_ral}
\input{sections/04_experiment}
\input{sections/05_conclusion}



%
\newpage
\bibliography{reference}
\bibliographystyle{abbrvnat}
%



\end{document}

%% file: preamble.tex
\usepackage{kantlipsum, lipsum}
\usepackage{dsfont}
\usepackage{multirow}
\usepackage{array}
\usepackage{amssymb} %
\usepackage{pifont} %
\usepackage{makecell}
\usepackage{algorithm}
\usepackage{algpseudocode}
\usepackage{amsmath}
\usepackage{multicol}
\usepackage{float}
\usepackage{siunitx}
\usepackage{tabularx}
\usepackage{booktabs}

\usepackage{xcolor}
\definecolor{GoogleBlue}{HTML}{1A73E8}
\definecolor{GoogleBlueLight}{HTML}{e8f0fe}
\definecolor{GoogleBlueDark}{HTML}{174ea6}
\definecolor{GoogleRed}{HTML}{D93025}
\definecolor{GoogleGray}{HTML}{767676}
\definecolor{figtextcolor}{HTML}{333333}
\definecolor{perceptioncol}{HTML}{1A73E8}
\definecolor{modelingcol}{HTML}{925BCF}
\definecolor{manipulationcol}{HTML}{C440A7}
\definecolor{reasoningcol}{HTML}{DA2F77}

\usepackage{subcaption}
\usepackage{opensans}
\DeclareCaptionFont{subfigcapsize}{\fontsize{8pt}{8pt}\selectfont}  %
\usepackage{xurl}
\usepackage[colorlinks=true, allcolors=GoogleBlue]{hyperref}
\usepackage[capitalise]{cleveref}
\crefname{section}{Sec.}{Secs.}  %
\crefname{subsection}{Sec.}{Secs.}  %
\crefname{subsubsection}{Sec.}{Secs.}  %
\crefname{appendix}{App.}{Apps.}  %

\usepackage{graphicx}
\graphicspath{{figures/}}

\usepackage[breakable]{tcolorbox}

\newcounter{takeaway}
\usepackage{tikz}
\usetikzlibrary{calc}
\newlength{\tikzwidth}
\newlength{\boxpad}
\newlength{\boxgap}
\newlength{\boxtextheight}

%% file: sections/00_abstract.tex
\begin{abstract}
Post-training with Reinforcement Learning (RL) has substantially improved reasoning in Large Language Models (LLMs) via test-time scaling. However, extending this paradigm to Multimodal LLMs (MLLMs) through verbose rationales yields limited gains for perception and can even degrade performance.

We propose Reinforced Attention Learning (\modelname), a policy-gradient framework that directly optimizes internal attention distributions rather than output token sequences. By shifting optimization from what to generate to where to attend, \modelname promotes effective information allocation and improved grounding in complex multimodal inputs. Experiments across diverse image and video benchmarks show consistent gains over GRPO and other baselines. We further introduce On-Policy Attention Distillation, demonstrating that transferring latent attention behaviors yields stronger cross-modal alignment than standard knowledge distillation. Our results position attention policies as a principled and general alternative for multimodal post-training.
\end{abstract}

%% file: sections/01_intro.tex
\begin{figure}[H]
    \centering
	\includegraphics[width=0.9\linewidth]{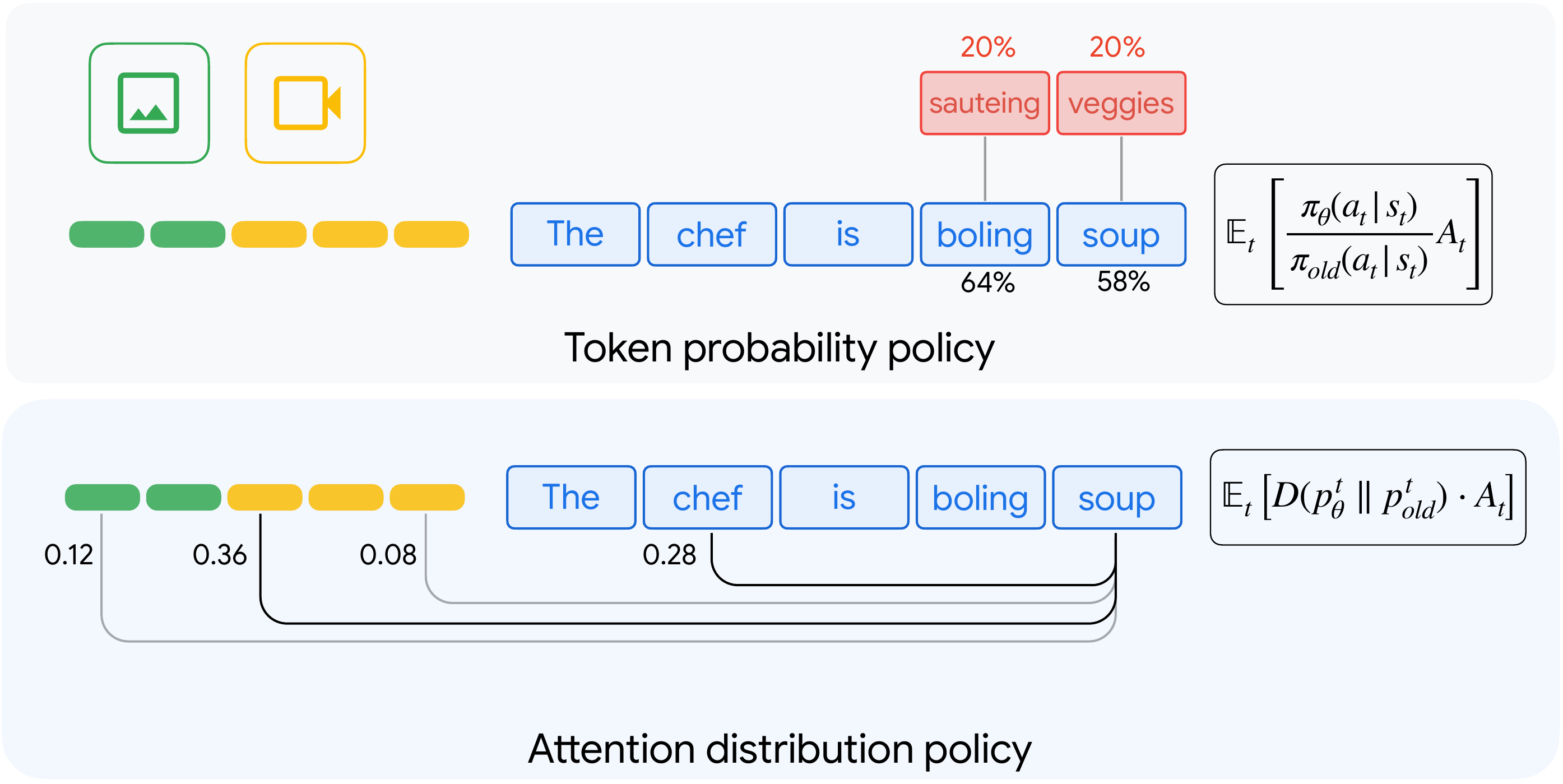}
    \caption{\textbf{\modelfullname formulates internal attention distributions as a policy.} Unlike traditional methods that optimize next-token probabilities (``what to generate''), our approach prioritizes the selective allocation of information (``where to focus''). By optimizing for the advantage, the model explores a high-reward attention policy that effectively isolates salient information from dense contexts.}
\label{fig:main}
\end{figure}

\section{Introduction}
Large Language Models (LLMs) have achieved remarkable proficiency in complex domains such as mathematics and programming. Beyond massive unsupervised pre-training \cite{gpt3, gemini}, post-training has emerged as a critical technology for eliciting long-form Chain-of-Thought (CoT) reasoning. Current paradigms predominantly employ Reinforcement Learning (RL) to optimize the model’s policy, utilizing rewards derived from learned models or programmatic verifiers to favor high-utility token sequences. Fundamentally, these policy gradient methods refine the next-token distribution to maximize expected rewards. Recent research has further highlighted a robust correlation between reasoning length and task accuracy—a phenomenon central to the concept of test-time scaling.

To extend these gains to multimodal LLMs (MLLMs), recent studies have attempted to incorporate ``thinking processes'' into Visual Question Answering (VQA) tasks. Under this paradigm, models are incentivized to generate exhaustive textual descriptions of visual inputs as a CoT prior to providing a final answer. However, our results reveal that for core perception tasks—such as image or video question answering—extensive textual reasoning provides only marginal gains and may even degrade the model's fundamental perceptual capabilities.

We attribute this limitation to the insufficiency of next-token prediction as the fundamental policy objective in MLLM post-training. In typical MLLM architectures, visual inputs are encoded as tokens and projected into the textual embedding space to serve as context for generation. Accurately answering fine-grained questions requires the model to precisely identify and attend to task-relevant visual information. This process is governed by the Transformer’s attention mechanism, which must learn to assign high weights to salient multimodal tokens. Standard RLHF, however, optimizes for the result (the token) rather than the process (the internal information allocation).

Inspired by this observation, we reformulate the post-training policy for MLLMs to operate directly on the attention distribution during generation. This yields \modelfullname (\modelname), an algorithm designed to optimize the model toward high-utility attention trajectories. Unlike traditional methods, \modelname treats the attention pattern itself as the policy: when a response receives a high reward, the algorithm encourages the underlying attention distribution by minimizing the divergence between the current and reference policies. Conversely, for low-reward responses, the model is penalized by increasing the divergence from those sub-optimal patterns. By shifting the optimization target from token likelihood to attention-based allocation, \modelname fine-tunes MLLMs more directly for multimodal alignment. Our results indicate that \modelname consistently outperforms Group Relative Policy Optimization (GRPO) across video and image benchmarks, particularly on perception-intensive tasks.

The efficacy of optimizing attention distributions naturally extends to \textbf{On-Policy Distillation}. While traditional distillation focuses on token-level probability alignment, we propose a dual-distillation approach that transfers knowledge via both token and attention distribution alignment. Our experiments indicate that the inclusion of attention distillation provides significant additional performance gains.

In summary, this paper introduces a novel post-training paradigm for MLLMs. Our contributions are as follows:

\begin{itemize}[left=0pt]
\item \modelfullname: We propose \modelname, a policy-gradient method that shifts the optimization objective from next-token prediction to attention-distribution alignment, enabling direct reinforcement of visual grounding rather than indirect supervision through textual outputs.

\item \textbf{On-Policy Attention Distillation}: We further extend this framework to an On-Policy Distillation setting, which substantially improves a student model’s ability to inherit fine-grained perceptual and grounding behaviors from a teacher.

\item \textbf{Empirical Validation}: Extensive experiments demonstrate that \modelname consistently improves upon GRPO across diverse visual question answering benchmarks that require fine-grained visual understanding and perception.

\end{itemize}

%% file: sections/02_related_works.tex
\section{Related Works}

\subsection{Post training LLMs through Reinforcement Learning}
Post-training is now the standard for aligning Large Language Models (LLMs) with human intent \cite{ouyang2022training}. The traditional RLHF pipeline involves Supervised Fine-Tuning (SFT), training a Reward Model (RM) to mimic human preferences, and optimizing the policy via Reinforcement Learning (RL) \cite{christiano2017deep}. While early RLHF methods significantly improved model safety and helpfulness \cite{bai2022constitutionalaiharmlessnessai}, they relied heavily on Proximal Policy Optimization (PPO) \cite{schulman2017proximal}. However, PPO’s actor-critic framework is memory-intensive due to the auxiliary critic model. Group Relative Policy Optimization (GRPO) \cite{deepseekmath2024} addresses this by replacing the critic with group-averaged reward estimates. This shift reduces computational overhead while maintaining high performance, particularly in verifiable domains like reasoning and code \cite{deepseekr12025}, leading to the domain of RL with Verifiable Rewards (RLVR).

Extending post-training to multimodal LLMs (MLLMs) introduces challenges beyond text-only alignment, including visual hallucination and robust cross-modal grounding \cite{liu2023llava,pope,zhu2023minigpt}. Recent methods adapt RLHF, RLVR, or Direct Preference Optimization (DPO) to improve visual grounding and reduce hallucinations \cite{qwen25vltechnicalreport,internvl3exploringadvancedtraining,visionr1}.

A persistent issue is \emph{modality bias}, where the model over-relies on linguistic priors or, conversely, overfits to superficial visual cues \cite{cai2025diagnosing}. To address this, recent work designs reward functions and training signals that discourage text-only shortcuts, penalize spurious visual correlations, and promote faithful, evidence-based responses \cite{xia2025visionary,papo}. 

Our approach targets the same goal by leveraging a fundamental information-selection mechanism: attention. Since cross-modal reasoning depends on identifying salient evidence in both modalities, directly shaping attention weights provides a principled way to control the cross-modal reasoning policy in our method, rather than relying solely on text-token-level policies.

\subsection{Distilling knowledge and beyond from teacher to student models}
Knowledge Distillation (KD) transfers knowledge from a high-capacity teacher to a student by matching softened output distributions rather than hard labels \cite{hinton2015distilling}. By providing richer supervisory signals, KD has been widely adopted for model compression, domain adaptation, and efficient deployment \cite{romero2015fitnetshintsdeepnets, zagoruyko2016paying}. In the context of large language models, distillation has been extended beyond output logits to intermediate representations, attention maps, and hidden states, enabling improved preservation of structural and reasoning behaviors \cite{sun2019patientknowledgedistillationbert, jiao-etal-2020-tinybert,wang2020minilm}.

More recent work has explored \emph{on-policy} distillation \cite{opd}, in which the student generates responses under its own policy and receives supervision from teacher evaluations along these trajectories. Compared to offline KD on static datasets, on-policy distillation mitigates exposure bias and better aligns the student’s generation distribution with deployment-time behavior. This paradigm is closely related to RL–based post-training, yet retains the stability and efficiency of supervised learning objectives.

In this work, we investigate knowledge distillation and attention distillation as alternative mechanisms for learning effective attention distributions. Incorporating attention-level supervision within an on-policy distillation framework provides a principled means of regularizing the student’s internal information allocation while maintaining policy alignment. This motivates our experimental study of on-policy attention distillation as a complementary approach to purely reward-driven optimization.

%% file: sections/03_ral.tex
\section{Reinforced Attention Learning}
Traditional reinforcement learning (RL) for Large Language Models (LLMs) typically optimizes a policy  by maximizing the expected return over the output token distribution. Modern off-policy algorithms like PPO and GRPO utilize a surrogate objective based on importance sampling:

\begin{equation}
\mathcal{L}_{\text{RL}} = \mathbb{E}_{t} \left[ \frac{\pi_\theta(a_t|s_t)}{\pi_{\text{old}}(a_t|s_t)} A_t \right]
\end{equation}
where $A_t$ denotes the advantage estimate. This formulation explicitly optimizes the \textbf{divergence between the current and the behavioral policy's token distributions}. While effective for maximizing immediate rewards, this token-level optimization often precipitates \textbf{diversity collapse}. In such cases, the model overfits to specific high-reward surface forms, thereby weakening its generalization across diverse reasoning patterns and leading to ``reward hacking'' of the linguistic structure rather than the underlying logic.

To mitigate this, we propose shifting the optimization target from the external output distribution to the \textbf{internal attention distributions}. By supervising how the model allocates its computational focus over contextual information, we provide a robust form of structural regularization. We frame the model's aggregate attention over an input prompt as an \textbf{information-gathering policy}.

\subsection{Aggregated causal Attention Distribution Policy}

We posit that the model’s internal attention mechanism constitutes an alternative, latent policy space. By reinforcing these internal distributions, we guide the model's reasoning process—specifically, how it integrates both the original context and its own generated rationale—without strictly constraining the output tokens to a narrow, brittle distribution.

Let the total sequence be $S = (x_1, \dots, x_T)$, where $x_1, \dots, x_P$ represents the prompt and $x_{P+1}, \dots, x_T$ represents the generated response. Let $\alpha_{t,i}$ represent the attention weight from the generated token at position $t$ to any preceding token at position $i$ ($i < t$), extracted from the final layer and averaged across all heads. For each generated token $t \in [P+1, T]$, we define the causal Attention Distribution Policy $p_{\theta}^t$ as the distribution over all preceding positions:

\begin{equation}
p_{\theta}^t(i) = \frac{\alpha_{t,i}}{\sum_{j=1}^{t-1} \alpha_{t,j}}, \quad \forall i \in [1, t-1]
\end{equation}

This formulation captures how the model attends to its own emerging reasoning in addition to the initial instructions and visual input.

\subsection{Advantage-Weighted Attention Divergence}

To ensure the model retains an effective \textbf{information-gathering policy}, we derive a per-token loss function. Drawing inspiration from the importance sampling ratios in PPO/GRPO, this objective encourages attention patterns that correlate with high rewards. The objective is defined as the expected divergence of the current attention policy and the old attention policy:

\begin{equation}
L_{\text{AttnRL}} = \mathbb{E}_{t} \left[ A_{t} \cdot D(p_{\theta}^t \parallel p_{\text{old}}^t) \right]
\end{equation}
where $A_t$ is the sequence-level advantage and $D(\cdot)$ is a symmetric, bounded divergence measure, such as Jensen-Shannon Divergence (JSD). Using JSD ensures training stability and behaves according to the sign of the advantage:

\begin{itemize}
    \item If $A_{t} > 0$, minimizing $L_{\text{AttnRL}}$ pulls the current policy $p_{\theta}$ toward the successful strategy $p_{\text{old}}$.
    \item If $A_{t} < 0$, the objective ``pushes'' $p_{\theta}$ away from the suboptimal strategy.
\end{itemize}

This per-token granularity prevents the ``vanishing gradient'' effect often encountered when averaging attention over long sequences, ensuring that even late-stage tokens in a long response are supervised by the internal attention objective.

\subsection{Combined Optimization Objective}

The final training objective integrates the standard token-level policy gradient with our internal attention regularizer:

\begin{equation}
\mathcal{L}_{\text{total}} = \mathcal{L}_{\text{RL}} + \lambda_{\text{attn}} L_{\text{AttnRL}}
\end{equation}
where $\lambda_{\text{attn}}$ is a hyperparameter balancing explicit output maximization with internal attention-level exploration. This dual-objective approach ensures the model remains linguistically flexible while maintaining a structured and reward-aligned reasoning process.

\subsection{Gradient Derivation}\label{ssec:ral_gradient}
The gradient of $\mathcal{L}_{\text{AttnRL}}$ with respect to the attention logits $e_{t,i}$ is derived via the chain rule. Let $J_t = \text{JSD}(p_\theta^t \| p_{\text{old}}^t)$.

\paragraph{Gradient w.r.t. Distribution.} The partial derivative of JSD with respect to the current distribution $p_\theta^t$ is:

$$\nabla_{p_\theta^t} J_t = \frac{1}{2} \ln \left( \frac{2 p_\theta^t}{p_\theta^t + p_{\text{old}}^t} \right)$$

\paragraph{Gradient w.r.t. Logits.} Using the softmax Jacobian $\frac{\partial p_j}{\partial e_i} = p_i(\delta_{ij} - p_j)$, the gradient for a specific logit $e_{t,i}$ is:

$$\nabla_{e_{t,i}} J_t = p_\theta^t(i) \left( \nabla_{p_\theta^t(i)} J_t - \sum_j p_\theta^t(j) \nabla_{p_\theta^t(j)} J_t \right)$$

\paragraph{Total Parameter Update.} The final gradient for the attention loss is the advantage-weighted accumulation across the sequence:

$$\nabla_\theta \mathcal{L}_{\text{AttnRL}} = \mathbb{E}_{\tau} \left[ A_\tau \sum_{t=P+1}^{T} \sum_{i=1}^{t-1} \left( \nabla_{e_{t,i}} J_t \right) \nabla_\theta e_{t,i} \right]$$

When $A_\tau > 0$, the update minimizes divergence from successful patterns. When $A_\tau < 0$, it pushes the model to explore alternative attention fragments, penalizing the specific reasoning path that led to a low reward.

\subsection{On-Policy Attention Distillation}

Beyond reward-driven optimization, the attention policy framework can be extended to an on-policy distillation setting where the student model $\pi_\theta$ aims to inherit the attention distribution, as the structural reasoning patterns, from a teacher model $\pi_\phi$. In this regime, we provide dense supervision by minimizing the divergence between the student's and teacher's internal attention distributions over the student's own generated trajectories.

\paragraph{Teacher-Student Alignment.} For each token $t$ in a trajectory $\tau$ sampled from the student policy $\pi_\theta$, we define the distillation loss as the sum of divergences between the student attention policy $p_\theta^t$ and the teacher attention policy $p_\phi^t$:

\begin{equation}
    \mathcal{L}_{\text{AttnDistill}} = \mathbb{E}{\tau \sim \pi_\theta} \left[ \sum_{t=P+1}^{T} \text{JSD}(p_{\theta}^t | p_{\phi}^t) \right]
\end{equation}

Crucially, this objective does not involve an advantage term $A_t$. The goal is pure structural imitation, ensuring that for any generated token, the student utilizes the same contextual evidence as the teacher. This provides a denser gradient signal than token-level KL divergence alone.

\paragraph{Unified Distillation Objective.} The final objective for on-policy distillation combines the RL objective, the generalized knowledge distillation (GKD) on output logits, and our proposed attention alignment:

\begin{equation}
\mathcal{L}_{\text{total}} = \mathcal{L}_{\text{RL}} + \mu \mathcal{L}_{\text{GKD}} + \gamma_{\text{attn}} \mathcal{L}_{\text{AttnDistill}}
\end{equation}
where $\mathcal{L}_{\text{GKD}}$ typically represents the reverse Kullback-Leibler divergence between the output distributions $\pi_\theta(a_t|s_t)$ and $\pi_\phi(a_t|s_t)$. $\mu$ controls the strength of the distillation and $\gamma_{\text{attn}}$ balances the strength of the attention distillation.

\paragraph{Gradient Flow.} The gradient of $\mathcal{L}_{\text{AttnDistill}}$ with respect to student parameters $\theta$ is derived similarly to \ref{ssec:ral_gradient}, but is purely aligning toward the teacher's distribution:

\begin{equation}
\nabla_\theta \mathcal{L}_{\text{AttnDistill}} = \mathbb{E}_{\tau \sim \pi_\theta} \left[ \sum_{t=P+1}^{T} \sum_{i=1}^{t-1} \left( \nabla_{e_{t,i}} J_t \right) \nabla_\theta e_{t,i} \right]
\end{equation}
where $J_t = \text{JSD}(p_{\theta}^t \| p_{\phi}^t)$. By sampling trajectories from the student ($\tau \sim \pi_\theta$), the model learns to maintain ``teacher-like'' attention patterns even when navigating states it would not encounter under the teacher's original distribution, thereby mitigating exposure bias in internal representations.

%% file: sections/04_experiment.tex
\section{Experiments}
\label{sec:experiments}

\begin{figure}[t]
    \centering
	\includegraphics[width=0.5\linewidth]{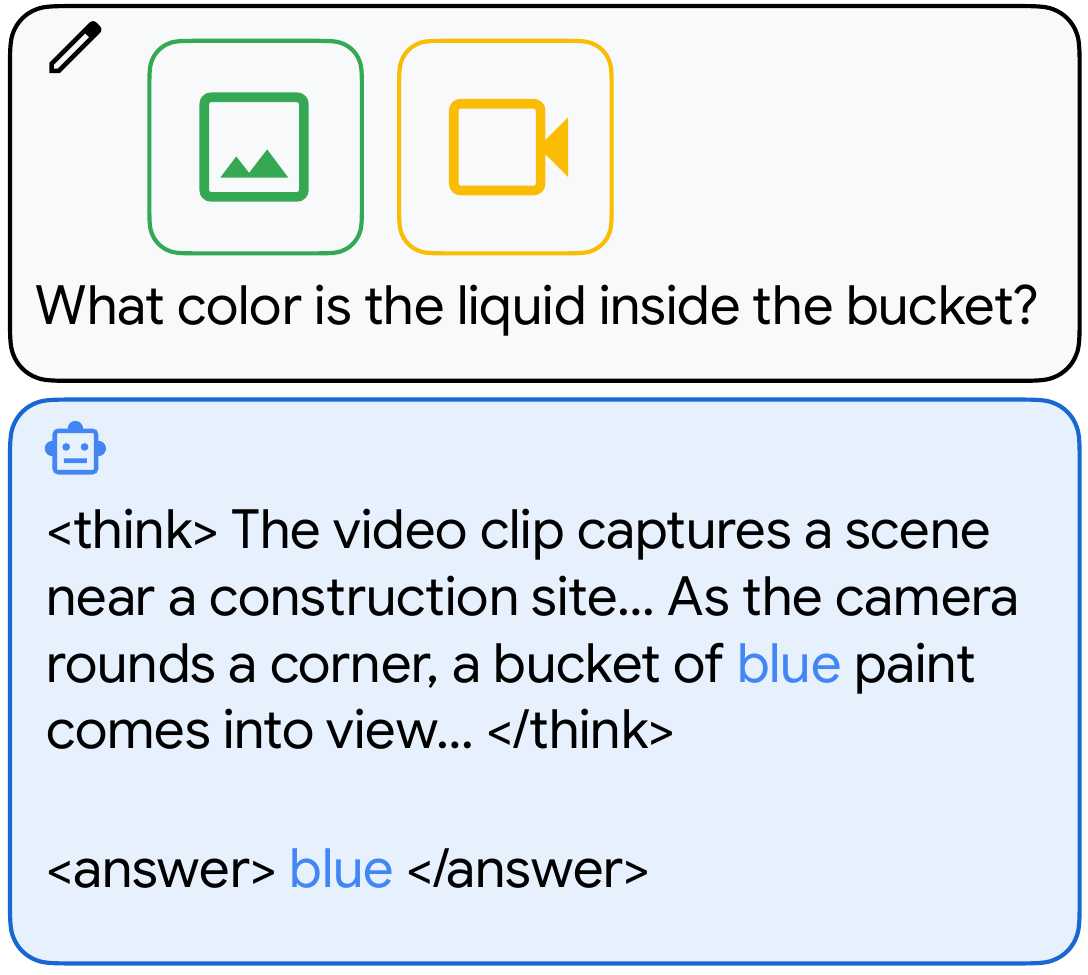}
    \caption{\textbf{Sample data of the SFT and RL training stages.} The SFT stage adapts the model to a ``think-and-answer'' paradigm, while the RL stage employs a reward function to verify the format and correctness of the rollout responses.}
\label{fig:data_sample}
\end{figure}

In this section, we empirically evaluate \modelname for post-training MLLMs. Our experiments investigate both the standard \modelname objective and the on-policy attention distillation variant, implemented on top of the Group Relative Policy Optimization (GRPO) framework.

\subsection{Experimental Setup}

\paragraph{Model Configurations} 
We adopt \textbf{Qwen-2.5-VL-7B} as our foundation MLLM. For the on-policy attention distillation experiments, we utilize \textbf{Qwen-2.5-VL-32B} as the teacher model. In all experimental settings, the visual encoder and multimodal projector are kept frozen, with gradients updated only for the language model backbone.

\paragraph{Training Pipeline}
Our training pipeline utilizes the \textsc{Video-R1}~\citep{videor1} dataset and consists of two primary stages, implemented using the \texttt{veRL} infrastructure~\citep{verl}:

\begin{itemize}
    \item \textbf{Supervised Fine-Tuning (SFT):} To align the model with our target reasoning schema, we perform SFT using the \textsc{Video-R1-COT-165k} subset. This dataset provides 165k instances of video-question pairs paired with structured Chain-of-Thought reasoning. Following the format in \Cref{fig:data_sample}, the ``thinking'' process and final answer are enclosed in <think> and <answer> tags, respectively. The computational overhead for this stage is roughly 10 hours on a cluster of 8$\times$ NVIDIA H100s.
    
    \item \textbf{Reinforcement Learning (RL):} We subsample the first 51.2k instances from the \textsc{Video-R1-260k} dataset and perform training for a single epoch. For each input, the policy generates $G=8$ rollouts to facilitate advantage estimation. This RL phase requires approximately 120 hours of compute on a cluster of 8$\times$ NVIDIA H100 GPU cluster.
\end{itemize}

\subsection{Implementation Details}

\paragraph{Visual Processing}
For the visual encoder, we set the maximum resolution of images to $5120 \times 28 \times 28$ pixels and the minimum to $128 \times 28 \times 28$ pixels. Video frames are sampled at 1 frame per second (fps), with the total number of frames capped at 128. Each individual frame is constrained to a maximum of $128 \times 28 \times 28$ pixels.

\begin{table}[h]
\centering
\begin{tabular}{@{}ll@{}}
\toprule
\textbf{Parameter} & \textbf{Value} \\ \midrule
Rollout batch size & 8 \\
KL Coefficient ($\beta$) & 0.04 \\
Sampling Temperature ($\tau$) & 1 \\
$\lambda_{attn}$ & $\{0.5, 1, 5\}$ \\
$\gamma_{attn}$ & $\{0.05, 0.5, 1\}$ \\ \bottomrule
\end{tabular}
\caption{Hyperparameters for \modelname experiments.}
\vspace{-4ex}
\label{tab:hyperparams}
\end{table}

\paragraph{Hyperparameters.}
We extract the attention weights of the last Transformer layer of the LLM by patching its attention implementation with eager attention. The attention weights are averaged across all attention heads. For RL training, we maintain a fixed sampling temperature of $\tau=0.9$ and a KL coefficient of $\beta=0.04$. The learning rate is set to $5\times 10^{-6}$ for SFT and $1\times 10^{-5}$ for RL. Detailed hyperparameter ranges for \modelname are provided in \Cref{tab:hyperparams}.

\paragraph{Baselines.} 
We primarily compare our method against the \textbf{Group Relative Policy Optimization (GRPO)} algorithm, which eliminates the need for a separate value model by computing the relative advantage of each response within a group of $G$ sampled outputs for the same query. For a given response $i$, the advantage $\hat{A}_{i}$ is calculated as:
\begin{equation}
    \hat{A}_{i} = \frac{r_i - \frac{1}{G} \sum_{j=1}^{G} r_j}{\text{std}(r_1, \dots, r_G)}
\end{equation}
where $r_i$ represents the total reward for response $i$. Our reward system is entirely rule-based, comprising two components:
\textbf{Accuracy Reward ($r_{acc}$):} We extract the content within the \texttt{<answer>...</answer>} tags and compare it against the ground truth. The model receives a reward of 1.0 for an exact match and 0.0 otherwise.
\textbf{Format Reward ($r_{fmt}$):} A regular expression-based verifier checks if the response strictly adheres to the template: \texttt{<think>...</think><answer>...</answer>}. A reward of 1.0 is granted for perfect formatting and 0.0 otherwise.
The final reward $r_i$ is a weighted combination: $r_i = 0.9 \cdot r_{acc} + 0.1 \cdot r_{fmt}$. These settings remain consistent across all experiments, including \modelname, on-policy attention distillation, the baseline GRPO and the GRPO-based on-policy knowledge distillation.

We also include Video-R1-7B for comparison, which was introduced alongside the dataset. This model is trained using Temporal-GRPO, a proposed variant of the standard GRPO framework that incorporates a temporal coherence reward. Specifically, this method perturbs the chronological ordering of video frames and provides an auxiliary reward when the model demonstrates higher accuracy on correctly sequenced inputs compared to shuffled ones. Since Video-R1-7B utilizes the same Qwen-2.5-VL-7B backbone as our other configurations, it facilitates a strictly controlled and fair comparison across our evaluation suite.

\subsection{Evaluation Benchmarks}
We evaluate \modelname across a broad spectrum of vision-centric benchmarks to assess spatial, temporal, and reasoning-intensive capabilities.

\paragraph{Image QA Tasks.} For static image understanding, we utilize:(1) $V^*$ Bench for fine-grained visual search and spatial reasoning;(2) MMMU Pro \citep{mmmupro} for multi-step expert-level knowledge;(3) MME \citep{mme} for basic perception and high-level cognition;(4) MuirBench \citep{muirbench} for robustness across diverse image types;(5) ChartQA \citep{chartqa} for complex data extraction;(6) VizWiz \citep{vizwiz} for real-world visual grounding;(7) Blink \citep{blink} for foundational perception tasks; and (8) CVBench \citep{cvbench} for benchmarking core computer vision capabilities. 

\paragraph{Video QA Tasks.} To evaluate temporal reasoning and long-context integration, we adopt:(1) LongVideoBench \citep{longvideobench} for long-range referring reasoning;(2) NExT-QA \citep{nextqa} for causal and temporal action explanation;(3) Video-MME \citep{videomme} for comprehensive multi-domain evaluation;(4) Video-MMMU \citep{videommmu} for expert-level knowledge acquisition;(5) LVBench \citep{lvbench} for extreme long-form comprehension;(6) MVBench \citep{mvbench} for multi-task temporal perception; and(7) TempCompass \citep{tempcompass} for sensitivity to motion and temporal order.

\input{tables/main_res_video}
\input{tables/main_res_image}

\subsection{Main Results}
\Cref{tab:general_mm_results} and Table~\ref{tab:video_bench_results} present a comprehensive comparison of \modelname against Group Relative Policy Optimization (GRPO) and standard on-policy distillation across a diverse suite of image and video VQA benchmarks. Overall, \modelname consistently surpasses both the GRPO baseline and the base Qwen-2.5-VL-7B model. These results suggest that optimizing internal attention distributions provides a more stable and effective learning signal for multimodal reasoning than token-level policy gradients alone.

\paragraph{Image VQA.} Under the reinforcement learning paradigm, \modelname outperforms GRPO across all eight image benchmarks. We observe significant improvements on $V^*$ (+5.8), MME (+94.1), ChartQA (+2.8), and VizWiz (+3.8). These gains indicate that attention-level supervision strengthens visual grounding and compositional reasoning, particularly for perception-intensive and document-based tasks. Crucially, \modelname not only mitigates the performance degradation often introduced by GRPO relative to the base model (e.g., on $V^*$ and VizWiz) but also exceeds the original Qwen-2.5-VL-7B performance across all metrics. This suggests that our approach fosters genuine generalization rather than narrow reward-model overfitting.

In the on-policy distillation regime, the integration of attention distillation yields superior results over standard distillation on 7 out of 8 benchmarks, with marked increases on $V^*$ (+3.6) and MuirBench (+1.8). These findings demonstrate that supervising where a teacher model attends provides a crucial, complementary signal that simple output imitation lacks. Notably, the strong performance on $V^*$, which specifically probes fine-grained object attributes and spatial relationships, underscores \modelname's efficacy in resolving complex scene geometries.

\paragraph{Video VQA.} On long-video benchmarks (Table~\ref{tab:video_bench_results}), \modelname outperforms GRPO on 6 out of 7 datasets. The most pronounced improvements occur on LongVideoBench (+2.2), NExTQA (+3.4), and MVBench (+1.5), all of which demand robust temporal understanding and multi-hop reasoning. While GRPO maintains a marginal lead on VideoMMMU, \modelname remains highly competitive, suggesting that attention-level optimization does not compromise domain-specific factual accuracy.

Similarly, under on-policy distillation, attention-enhanced alignment improves performance on every benchmark except for ties on LongVideoBench and MVBench. The substantial gains on NExTQA (+4.4) and VideoMME (+2.6) suggest that attention alignment is particularly potent for long-context temporal reasoning and precise event localization.

\paragraph{Implications.} Across both modalities and training regimes, attention-centric learning provides consistent and well-distributed performance increments. In contrast to GRPO, which can exhibit benchmark-specific trade-offs or degrade base model capabilities, \modelname delivers uniform gains. This supports our hypothesis that supervising internal information allocation offers a more stable and generalizable training signal than pure token-level gradients. Furthermore, the success of attention distillation confirms that attention distributions serve as a transferable and semantically rich representation of reasoning behavior.

In summary, these results validate attention-based policy optimization as a robust complement for, conventional RL and distillation in MLLM post-training, especially for perception-heavy and long-horizon video understanding.

\begin{figure*}[t]
    \centering
	\includegraphics[width=0.8\linewidth]{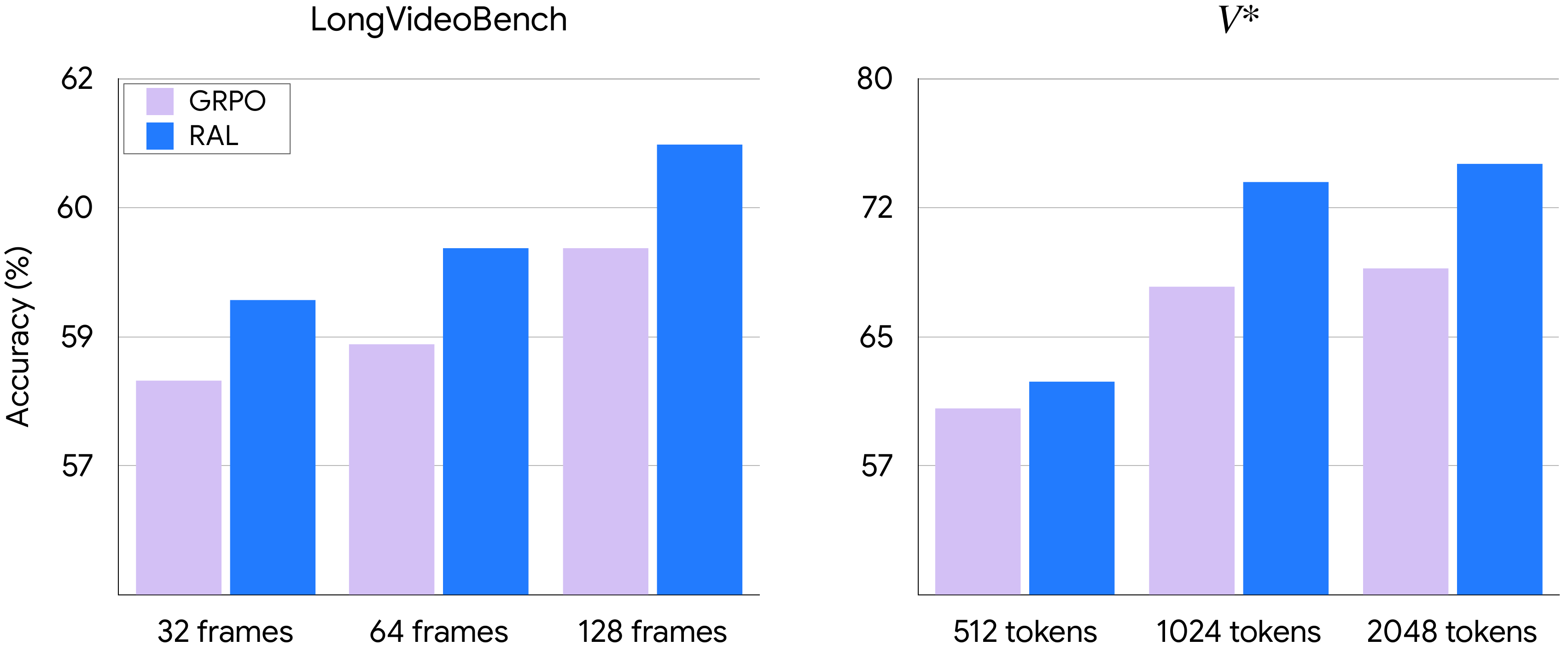}
    \caption{\textbf{\modelname improves GRPO along the increasing video frames or image resolution.}}
\label{fig:ablation_resolution}
\end{figure*}

\subsection{Ablation Studies}\label{ssec:ablation}
\paragraph{\modelname yields consistent improvements across varying visual resolutions and frame rates.} To investigate the robustness of these gains relative to visual information density, we analyze the performance of \modelname versus GRPO across different video sampling lengths and image resolutions. We utilize LongVideoBench and $V^*$ as our evaluation suite: the former requires extracting sparse, salient information from extremely long video sequences, while the latter features high-resolution images of complex scenes that demand fine-grained reasoning. Together, these benchmarks facilitate a diagnostic analysis of cross-modal reasoning capabilities across diverse temporal and spatial densities. 

We evaluate performance on LongVideoBench using maximum frame counts of 32, 64, and 128. For $V^*$, we vary the maximum token budget per image (512, 1024, and 2048 tokens). Here, a unit image patch is defined as a $28\times28$ square; thus, 1024 tokens correspond approximately to an $896\times896$ image. 

As illustrated in Figure~\ref{fig:ablation_resolution}, \modelname consistently outperforms GRPO across all temporal scales, demonstrating the efficacy of attention-based policies in locating salient cues within dense temporal contexts. On $V^*$, the performance margin widens as image resolution increases, rising from a +1.6 gain at 512 tokens to a significant +6.3 gain at 2048 tokens. This trend indicates that the advantages of \modelname become increasingly pronounced as visual information becomes more granular, suggesting superior scalability for high-fidelity multimodal understanding.

\paragraph{Is an Explicit Thinking Process Necessary for VQA?}
To isolate the impact of attention-level supervision, we investigate whether \modelname improves cross-modal reasoning even in the absence of an explicit thinking process. We introduce \modelname-zero, a variant where the thinking process is completely removed from both the SFT and RL stages. In this configuration, the model is trained to generate the final answer directly. By excluding high-reward text tokens (rationales), the training signal is dominated by the attention policy, allowing us to evaluate its intrinsic contribution to visual understanding.

We maintain all hyper-parameters and data volumes from the main experiments but modify the data format and reward functions. During SFT, the \texttt{\textless think\textgreater} blocks (see Figure~\ref{fig:data_sample}) are stripped, forcing the model to produce only the final response. Correspondingly, the RL stage employs a format reward that penalizes any output different from the \texttt{\textless answer\textgreater...\textless/answer\textgreater} structure.

As shown in Table~\ref{tab:video_bench_results}, \modelname-zero surpasses the base model on 5 out of 7 long-video benchmarks and outperforms the full GRPO baseline on 5 benchmarks. Notably, \modelname-zero achieves state-of-the-art performance on NExTQA (temporal reasoning), VideoMME (comprehensive video understanding), and LVBench (long-video event understanding). These results demonstrate that the attention policy space is significantly underexplored; \modelname effectively unlocks visual understanding capabilities by optimizing the distribution of internal attention weights.

On image-based benchmarks (Table~\ref{tab:general_mm_results}), \modelname-zero improves upon the base model in 4 out of 8 cases, achieving the highest scores among RL-based methods on MMMU-Pro and VizWiz. Given that MMMU-Pro tests complex visual knowledge and VizWiz focuses on fine-grained object recognition, these gains suggest that optimizing attention distributions serves as a powerful bridge between modalities. This confirms that policy gradient methods applied directly to attention mechanisms can induce superior cross-modal reasoning, even without explicit verbalized logic.

%% file: tables/main_res_video.tex

\begin{table*}[t]
\centering
\resizebox{\linewidth}{!}{
    \begin{tabular}{c | c c c c c c c}
    \Xhline{3\arrayrulewidth} 
    Method                         & LongVideoBench & NExTQA & VideoMME & VideoMMMU & LVBench & MVBench & TempCompass \\
    \midrule
    Qwen-2.5-VL-7B                 & 57.0      & 73.7       & 61.6      & 47.6      & 40.5      & 63.4      & 69.5 \\
    \hline
    \multicolumn{8}{c}{\textit{Reinforcement Learning}} \\
    \hline
    Video-R1                       & 56.5      & 65.3       & 62.5      & 44.7      & 43.5      & 62.0      & 68.0 \\
    GRPO                           & 57.9      & 70.7       & 62.0      & \textbf{49.7} & 43.9          & 64.0          & 68.3 \\
    \colorbox{googleblue!40}{\modelname} & \textbf{60.1} & \textbf{74.1} & \textbf{63.4} & 48.6          & \textbf{44.2} & \textbf{65.5} & \textbf{70.0} \\
    \colorbox{googlegreen!40}{\modelname-zero}                & 58.8    & 76.2  & 65.1  & 49.2   & 45.9     & 62.7      & 68.5\\
    \hline
    \multicolumn{8}{c}{\textit{On-Policy Distillation}} \\
    \hline
    Kowledge Distillation          & 59.7          & 70.9          & 61.3          & 47.3          & 43.7          & 65.5          & 69.2 \\
    \colorbox{googleblue!40}{+ Attention Distillation} 
                                   & \textbf{59.7} & \textbf{75.3} & \textbf{63.9} & \textbf{48.5} & \textbf{44.7} & \textbf{65.5} & \textbf{70.3} \\
    \bottomrule
    \end{tabular}%
}
\caption{\textbf{Performance comparison across long-video benchmarks.} We evaluate \modelname and its variants against baselines on tasks requiring extended visual context. Results for \colorbox{googleblue!40}{\modelname} and \colorbox{googleblue!40}{Attention Distillation} are highlighted in blue. \colorbox{googlegreen!40}{\modelname-zero} (discussed in Section \ref{ssec:ablation}) denotes the variant where explicit thinking process are removed, isolating the impact of the attention-based policy gradient. These results demonstrate that optimizing internal attention distributions provides a robust, complementary signal to token-based reinforcement learning, consistently enhancing multimodal understanding.}
\label{tab:video_bench_results}
\end{table*}

%% file: tables/main_res_image.tex

\begin{table*}[t]
\centering
\resizebox{\linewidth}{!}{
    \begin{tabular}{c | c c c c c c c c}
    \Xhline{3\arrayrulewidth} 
    Method                          & Vstar & MMMUpro & MME     & MuirBench & ChartQA & VizWiz & Blink & CVBench \\
    \midrule
    Qwen-2.5-VL-7B                  & 70.7  & 36.4    & 2309.3  & 44.9      & 84.0    & 71.2   & 56.3  & 77.6   \\
    \hline
    \multicolumn{9}{c}{\textit{Reinforcement Learning}} \\
    \hline
    Video-R1                        & 66.5  & 36.0    & 2266.2  & 40.9      & 84.2    & 63.5    & 47.4  &   72.3\\
    GRPO                            & 68.6            & 36.8          & 2258.7          & 43.9          & 81.7          & 67.9            & 54.9           & 78.1   \\
    \colorbox{googleblue!40}{\modelname}  & \textbf{73.3}   & \textbf{37.8} & \textbf{2352.8} & \textbf{47.4} & \textbf{86.4} & \textbf{71.7}   & \textbf{57.2}  & \textbf{79.0} \\
    \colorbox{googlegreen!40}{\modelname-zero}      & 72.3          & 38.4          & 2306.2          & 43.4          & 82.7          & 71.9          & 55.4 & 78.4 \\
    \hline
    \multicolumn{9}{c}{\textit{On-Policy Distillation}} \\
    \hline
    On-policy Distillation          & 68.1          & 37.8          & 2344.7          & 39.9          & 81.7          & 71.2          & 56.3 & \textbf{78.7} \\
    \colorbox{googleblue!40}{+ Attention Distillation} 
                                    & \textbf{72.3} & \textbf{38.4} & \textbf{2345.1} & \textbf{43.4} & \textbf{83.6} & \textbf{72.1} & \textbf{57.2} & 78.4 \\
    \bottomrule
    \end{tabular}%
}
\caption{\textbf{Performance comparison across image VQA benchmarks.}}
\label{tab:general_mm_results}
\end{table*}

%% file: sections/05_conclusion.tex
\section{Conclusion}
We introduced \modelfullname, a MLLM post-training paradigm that shifts optimization from text token distribution to internal attention distributions. By treating attention as a policy, \modelname directly reinforces visual grounding and perceptual focus, addressing a fundamental limitation of outcome-based RL methods that neglect the underlying cross-modal reasoning process.

Our experiments across diverse image and long-video benchmarks demonstrate that \modelname consistently outperforms the base Qwen-2.5-VL-7B and GRPO baselines. Notably, our approach provides more stable and uniform gains than token-level RL, which can occasionally degrade base model performance. These results validate our hypothesis that supervising internal information allocation yields a more reliable and generalizable training signal than next-token gradients alone. Furthermore, we showed that this attention-centric perspective extends naturally to on-policy distillation, where transferring ``where to focus'' provides a complementary and semantically rich signal that surpasses simple output imitation.

Ultimately, this work establishes attention distributions as a first-class optimization target for multimodal alignment. By reinforcing internal computation pathways, \modelname offers a principled, process-aware alternative to standard RLHF. We believe this perspective paves the way for future research into fine-grained credit assignment and the optimization of other internal structures, such as MoE routing or cross-modal fusion, to foster more robust and grounded multimodal intelligence.